\PassOptionsToPackage{dvipsnames}{xcolor}
\documentclass{article} % For LaTeX2e
\usepackage[final]{colm2025_conference}

\usepackage{microtype}
\usepackage{hyperref}
\usepackage{url}
\usepackage{booktabs}

\usepackage{lineno}

\definecolor{darkblue}{rgb}{0, 0, 0.5}
\hypersetup{colorlinks=true, citecolor=darkblue, linkcolor=darkblue, urlcolor=darkblue}

\usepackage[T1]{fontenc}
\usepackage{wrapfig}
\usepackage{colortbl}
\usepackage{textcomp}
\usepackage{siunitx}
\usepackage[font=small]{caption}
\usepackage{subcaption}
\usepackage{graphicx}
\usepackage[most]{tcolorbox}
\tcbuselibrary{fitting}
\tcbuselibrary{raster}
\usepackage{amsmath,amssymb}
\usepackage{booktabs}
\usepackage{multirow}
\usepackage{makecell}
\usepackage{titlesec}
\usepackage{fdsymbol}

\titlespacing*{\section}{0pt}{0.60ex plus 0.2ex minus 0.2ex}{0.60ex plus 0.2ex minus 0.2ex}
\titlespacing*{\subsection}{0pt}{0.60ex plus 0.2ex minus 0.2ex}{0.60ex plus 0.2ex minus 0.2ex}
\titlespacing*{\subsubsection}{0pt}{0.60ex plus 0.2ex minus 0.2ex}{0.60ex plus 0.2ex minus 0.2ex}
\usepackage[normalem]{ulem}
\useunder{\uline}{\ul}{}
\usepackage{enumitem}
\setlist[itemize]{noitemsep}
\usepackage{pgfplots}
\pgfplotsset{width=10cm,compat=1.9}
\usetikzlibrary{calc} 
\usepgfplotslibrary{external}
\makeatletter
\def\adl@drawiv#1#2#3{%
        \hskip.5\tabcolsep
        \xleaders#3{#2.5\@tempdimb #1{1}#2.5\@tempdimb}%
                #2\z@ plus1fil minus1fil\relax
        \hskip.5\tabcolsep}
\newcommand{\cdashlinelr}[1]{%
  \noalign{\vskip\aboverulesep
           \global\let\@dashdrawstore\adl@draw
           \global\let\adl@draw\adl@drawiv}
  \cdashline{#1}
  \noalign{\global\let\adl@draw\@dashdrawstore
           \vskip\belowrulesep}}
\makeatother

\DeclareMathOperator{\E}{\mathbb{E}}
\DeclareMathOperator*{\argmax}{arg\,max}

\newcommand*{\img}[1]{%
    \raisebox{-.3\baselineskip}{%
        \includegraphics[
        height=\baselineskip,
        width=\baselineskip,
        keepaspectratio,
        ]{#1}%
    }%
}

\title{Optimizing LLM-Based Multi-Agent System with Textual Feedback: A Case Study on Software Development}

\author{Ming Shen\textsuperscript{$\varheartsuit$}\thanks{Work done during internship at Amazon}\hspace{5pt}, Raphael Shu\textsuperscript{$\vardiamondsuit$}, Anurag Pratik\textsuperscript{$\vardiamondsuit$}, James Gung\textsuperscript{$\vardiamondsuit$}, Yubin Ge\textsuperscript{$\vardiamondsuit$}, \\ \textbf{Monica Sunkara\textsuperscript{$\vardiamondsuit$}, Yi Zhang\textsuperscript{$\vardiamondsuit$}} \\
\textsuperscript{$\varheartsuit$}Arizona State University \textsuperscript{$\vardiamondsuit$}Amazon Web Services
}

\begin{document}

\ifcolmsubmission
\linenumbers
\fi

\maketitle

% \begin{verbatim}
%    \usepackage[dvips]{graphicx} ...
%    \includegraphics[width=0.8\linewidth]{myfile.eps}
% \end{verbatim}
% or % Apr 2009 addition
% \begin{verbatim}
%    \usepackage[pdftex]{graphicx} ...
%    \includegraphics[width=0.8\linewidth]{myfile.pdf}
% \end{verbatim}

\begin{abstract}

We have seen remarkable progress in large language models (LLMs) empowered multi-agent systems solving complex tasks necessitating cooperation among experts with diverse skills. However, optimizing LLM-based multi-agent systems remains challenging. 
In this work, we perform an empirical case study on group optimization of role-based multi-agent systems utilizing natural language feedback for challenging software development tasks under various evaluation dimensions.
We propose a two-step agent prompts optimization pipeline: identifying underperforming agents with their failure explanations utilizing textual feedback and then optimizing system prompts of identified agents utilizing failure explanations. 
We then study the impact of various optimization settings on system performance with two comparison groups: online against offline optimization and individual against group optimization. For group optimization, we study two prompting strategies: one-pass and multi-pass prompting optimizations.
Overall, we demonstrate the effectiveness of our optimization method for role-based multi-agent systems tackling software development tasks evaluated on diverse evaluation dimensions, and we investigate the impact of diverse optimization settings on group behaviors of the multi-agent systems to provide practical insights for future development.

\end{abstract}

\section{Introduction}

Autonomous agents utilizing large language models have achieved promising results on tasks across various domains such as reasoning \cite{yao2023react, shinn2023reflexion, ge-etal-2025-tremu}, code generation \cite{shinn2023reflexion}, tool usage \cite{cai2023large}, and embodied AI \cite{Ahn2022DoAI}. 
Recent studies have demonstrated that incorporating synergistic agents into multi-agent collaboration frameworks substantially further enhances NLP problem-solving abilities \cite{zhuge2023mindstorms, Liang2023EncouragingDT, du2024improving, wu2024mathchat, wang-etal-2024-unleashing, li2023camel, Hao2023ChatLLMNM, zhang2024building, Dong2023SelfcollaborationCG, jiang-etal-2023-llm, wu2024autogen, chen2024agentverse} and successfully addresses complex real-world challenges, including human behavior simulation \cite{Park2023GenerativeAI}, software development \cite{qian-etal-2024-chatdev, hong2024metagpt}, code issue resolving \cite{Tao2024MAGISLM}, and web task execution \cite{zhang2024webpilot}. In our study, we focus on role-based multi-agent systems, where LLM-based agents are assigned distinct roles to accomplish a task collaboratively. 

A primary constraint inherent to LLMs is their reliance on prompt design \cite{NEURIPS2020_1457c0d6, 10.1145/3411763.3451760, si2023prompting}, which also extends to LLM-based agents. 
% which are the fundamental elements of any multi-agent framework utilizing LLMs. 
Specifically, it is crucial for developers to meticulously curate agent prompts, considering their respective roles and responsibilities in the collaboration framework. 
% To this end, various automatic agent prompt optimization methods have been proposed in recent works. 
To this end, various automatic agent prompt optimization methods have been proposed in recent works, such as DSPy \cite{khattab2024dspy}, GPTSwarm \cite{zhuge2024gptswarm}, and TextGrad \cite{Yuksekgonul2024TextGradA}. 
% DSPy \cite{khattab2024dspy} optimize LLM prompts via random search and bootstrapping.
% Analogous to our research, GPTSwarm \cite{zhuge2024gptswarm} employs Language Learning Models (LLMs) as optimizers for refining agent prompts, while TextGrad \cite{Yuksekgonul2024TextGradA} enhances agent prompts by leveraging natural language feedback.
However, none of these works have conclusively demonstrated the efficacy of their approaches within a genuine role-based multi-agent system, wherein agents take on various roles to tackle a complex task collaboratively. 
Furthermore, current works primarily focus on traditional NLP benchmarks such as MMLU \cite{hendrycks2021measuringmmlu}, MATH \cite{hendrycks2021measuringmath}, and HumanEval \cite{chen2021evaluating}. In contrast, our study aims to address the more intricate real-world software development task. 
% More detailed rationales behind selecting software development as our case study are addressed in \S\ref{sec:difficulty}. 

In this work, we propose a two-step agent prompt optimization framework utilizing natural language feedback of the multi-agent system along various evaluation criteria for software development tasks (\S\ref{sec:optimization_pipeline}). 
% ; however, our work is distinguished by the increased complexity of our multi-agent setting. 
In the first step, we employ an LLM-based locator to pinpoint underperforming agents, taking into account their roles and natural language feedback of the multi-agent system. The locator also provides fine-grained explanations for their underperformance. 
In the second step, we utilize an LLM-based optimizer \cite{zhou2023large, pryzant-etal-2023-automatic, yang2024large} to optimize the system prompts of identified underperforming agents based on fine-grained explanations. 
We demonstrate the efficacy of our proposed optimization pipeline across various evaluation dimensions pertinent to software development tasks. For each evaluation dimension, we collect natural language feedback using either model-based or rule-based methods.

Considering the inherent complexity of the multi-agent system, we further explore the impact of different optimization settings on the multi-agent system's performance (\S\ref{sec:optimization_setting}). Concretely, we investigate two comparison groups: online against offline optimization and individual against group optimization. In the online setting, agents interact with the environment to collect feedback during optimization; however, feedback is collected beforehand in the offline setting. In the individual optimization setting, one agent is optimized at a time during each optimization step, and in the group setting, all agents are optimized in each step. For group optimization, we further investigate two prompting strategies during the optimization step: one-pass and multi-pass prompting. In one-pass group optimization, agents are optimized together with one inference pass, whereas in multi-pass group optimization, agents are optimized with separate inference passes.
We show that online and offline optimizations are both effective, although offline performs slightly worse than online. We then show that optimizing all agents at each optimization step is necessary for our pipeline to outperform baselines consistently. 
Finally, we don't observe apparent performance differences between one-pass and multi-pass prompting, so one could choose one-pass prompting optimization for efficiency.

% The contributions of our work can be summarized as follows:
The contributions of our work are as follows:
\begin{itemize}
    \item We investigate optimizing a role-based multi-agent system on a complex real-world problem - software development. We show that the LLM-based multi-agent system can be effectively optimized utilizing textual feedback.
    \item We study the software development optimization problem along five distinct dimensions, whereas existing literature mainly focuses on a single evaluation metric for simpler tasks.
    \item We propose a two-step optimization pipeline by first locating and then optimizing underperforming agents. Experiment results show that our proposed pipeline outperforms baselines along the evaluation dimensions we study.
    \item We compare optimization effectiveness in various settings, including online against offline and individual against group optimization. We demonstrate online and group optimizations as more effective settings than individual and offline.
    % We then show that group optimization is equally effective as separate optimization but more efficient with fewer API calls.

\end{itemize}

\section{Related Works}

\subsection{LLM-Based Agents}

LLM-based agents \cite{yao2023react, shinn2023reflexion} have achieved impressive results on various NLP tasks. Applications \cite{autogpt} utilizing a single LLM-based agent, such as AutoGPT \cite{autogpt} and LangChain \cite{langchain}, can accomplish more functionalities and provide opportunities to a diverse audience spectrum, including developers and even non-technical users.
More recently, collaborations among multiple LLM-based agents demonstrated even more capabilities. 
\citet{Liang2023EncouragingDT} and \citet{du2024improving} incorporate multiple LLM-based agents into a debate framework, aiming to stimulate divergent thinking and enhance both factuality and reasoning capabilities.
Utilizing the unique strengths and knowledge of each individual agent, LLM-based agents can also be synergistically integrated to collaboratively enhance problem-solving capabilities on a broad range of tasks \cite{Hao2023ChatLLMNM, zhuge2023mindstorms, zhang2024building, ge-etal-2023-ask, Dong2023SelfcollaborationCG, wu2024mathchat, zhang2024webpilot}. In our work, we focus on role-based multi-agent system \cite{wang-etal-2024-unleashing, li2023camel, Park2023GenerativeAI, qian-etal-2024-chatdev, wu2024autogen, Tao2024MAGISLM, chen2024agentverse}, in which individual agent components are assigned distinct roles to interact and collaborate to accomplish a task effectively. Among the above role-based multi-agent systems, \citet{hong2024metagpt} and \citet{qian-etal-2024-chatdev} design agent roles inspired by Standardized Operating Procedures (SOPs) to solve software development tasks collaboratively.

\subsection{Prompt Optimization}
\label{sec:related_works_prompt_optim}

Studies have demonstrated the critical role of prompt engineering in unlocking the potential of LLMs \cite{NEURIPS2020_1457c0d6, gao-etal-2021-making, wei2022chain, 10.1145/3560815}. However, manually curating prompts with human effort is both time-consuming and expensive. Hence, investigating methods to perform prompt engineering automatically becomes a viable solution \cite{shin-etal-2020-autoprompt, deng-etal-2022-rlprompt, prasad-etal-2023-grips}. 
In our work, we adopt the recent trend of adopting LLMs as optimizers to optimize system prompts of the underperforming agents identified in the locator step.
As summarized by \citet{Ma2024AreLL}, recent works focusing on LLMs as optimizers can be classified into two categories: resampling-based and reflection-based.
Resampling-based methods \cite{zhou2023large, li-etal-2023-robust} sample around the current best prompts for better prompt candidate generation using LLMs while keeping the semantic meanings. On the other hand, reflection-based methods \cite{pryzant-etal-2023-automatic, Sun2023AutoHintAP, ye-etal-2024-prompt, wang2024promptagent, yang2024large, guo2024connecting} explicitly or implicitly leverage feedback or historical information to refine current prompts. Our optimization step falls in the reflection-based category.

\subsection{Agent Optimization}

We have seen various prompt optimization methods in \S\ref{sec:related_works_prompt_optim}, and it is also a critical topic for agent optimization. DSPy \cite{khattab2024dspy} views LLM-based systems as programs and proposes to build and optimize them in a programmatic fashion. DSPy can optimize LLM inference prompts, including few-shot examples and system instructions, through search algorithms in a combinatorial space. TextGrad \cite{Yuksekgonul2024TextGradA} takes a different perspective of backpropagation being a general and powerful framework to optimize the LLM-based agent system based on natural language feedback. However, none of them proves the effectiveness of their methods in a role-based multi-agent system setting through their experimental results. 
Research works in another agent optimization direction, such as DyLAN \cite{liu2024a} and GPTSwarm \cite{zhuge2024gptswarm}, focus on optimizing the workflow of multi-agent systems. They usually view multi-agent systems as directed graphs and optimize the graph by selecting or pruning existing edges that represent information exchange between agents. However, a fundamental limitation of such an optimization paradigm is that it relies on the assumption of the multi-agent system being decentralized, where all the agent components can draw conclusions regarding a given task, and later, the final output is generated using a consensus schema such as majority voting. 
This optimization paradigm is less applicable in scenarios where information flows are predefined or where agent components assume varied responsibilities to accomplish a task collaboratively, particularly in the context of real-world tasks.
Finally, a significant difference between our work and all the above works is that other than traditional NLP tasks, we study optimization for software development tasks, which are complex, open-ended tasks.

\section{Software Developemt Task}

In our work, we choose software development as the case study task for optimizing LLM-based multi-agent systems. Software development tasks require complete software solution code based on detailed software requirement descriptions spanning a broad
range of applications, such as board games and social networking. 

\subsection{Why Software Development?}
\label{sec:difficulty}
The reasons we choose software development tasks are three-fold. 
% First, unlike traditional NLP tasks such as natural language understanding \cite{hendrycks2021measuring}, mathematical reasoning \cite{cobbe2021training}, or single code function generation \cite{chen2021evaluating}, 
First, unlike traditional NLP tasks such as natural language understanding and mathematical reasoning, software development is a more suitable test bed for role-based multi-agent collaboration since it necessitates cooperation among multiple agents with diverse skills. For example, \cite{hong2024metagpt} designs a product manager for natural language description understanding, engineers for code reasoning, and code reviewers for code revision to tackle software engineer tasks collaboratively. 
Second, given the difficulty of evaluating solution code for open-ended software descriptions, there is generally no annotated ground-truth solution and standard evaluation metrics, unlike traditional NLP tasks. However, this gives us the flexibility to demonstrate the effectiveness of our optimization pipeline along various user-defined evaluation dimensions.
% As a result, unlike traditional NLP benchmarks where standard evaluation metrics are predefined, such as accuracy for multiple-choice questions and pass@1 for single code function generation, what dimension to optimize for software development is defined by user interest. This enables exploring optimization along various evaluation dimensions more easily.
Third, although previous works explored LLM-based agent optimization, group optimization of agents in a complex role-based multi-agent collaboration environment for real-world software development tasks has never been addressed.
% The challenges of our study include the difficulty of group optimization of multiple agents and the complexity and open-ended nature of the task itself.

\subsection{Optimization Dimensions}
\label{sec:optimization_dimensions}
% We explore group optimization of multi-agent system for software development along various user-defined evaluation dimensions in our study. For each dimension, a scalar value to directly assess the software solution code and feedback in natural language explaining the scalar score are generated. We generate scores and feedback using two high-level methodologies. First is rule-based method, where scores or feedback are generated based on heuristic rules or external tools. The second is model-based method, following recent works of LLMs as judges \cite{zheng2023judging, dubois2023alpacafarm, mcaleese2024llm}, where we guide large language models to evaluate solution code by carefully designed prompts. Next, we explain in detail the concrete methods we use for each dimension.
In our study, we explore optimization along five evaluation dimensions pertinent to software development tasks. In this section, we first introduce these five dimensions.

\begin{itemize}
  \item \textbf{Functionality} -- Functionality is the most crucial criterion for judging the quality of software code. It judges whether the software code meets all the requirements and specifications outlined in the software description. 
  \item \textbf{Robustness} -- Through the robustness dimension, we aim to measure whether the software is reliable enough to handle various unexpected user inputs or exceptions. 
  \item \textbf{Test Case Coverage} -- Test cases are essential for software code. To improve code quality and coverage, they help verify and validate whether the code is functioning as expected and identify bugs and errors in the code. We aim to optimize the software code to contain test cases to cover all aspects of the task description. 
  \item \textbf{Documentation} -- To make it easier for developers to understand, maintain, and collaborate on the solution code, we aim to optimize the system to generate enough proper documentation, such as comments and docstrings. 
  \item \textbf{Code Style Violation} -- Finally, to enhance code readability and consistency, we aim to optimize the software code to follow PEP 8 \footnote{\url{https://peps.python.org/pep-0008/}}, a style guide for Python code. 
\end{itemize}

\section{Group Optimization of LLM-Based Multi-Agent System with Feedback}

\subsection{Problem Formulation}
\label{sec:formulation}
We model an LLM-based multi-agent system as a directed graph $G = \{N, E\}$, where $N$ is a set of nodes and $E$ is a set of edges that are ordered pairs of nodes $E \subseteq \{(u, v) | (u, v) \in N \times N, u \neq v\}$. 
Each node $n \in N$ is a $LLM$ agent that takes in natural language form input $I_n$ combined with its agent prompt $P_n$ to generate a response $O_n = LLM(I_n \oplus P_n)$, where $\oplus$ stands for string combination. For role-based multi-agent systems, the role of agent $n$ is reflected in its prompt $P_n$. 
The prompts of all agent nodes in the graph form prompt group $\mathcal{P} = \{P_n | n \in N\}$.
An edge $(u, v) \in E$ represents that agent $u$ sends its response output $O_u$ to agent $v$. Suppose all the antecedent nodes for agent node $n$ are $A(n) = \{m \in N | (m, n) \in E\}$, the input $I_n$ for agent $n$ consists of all outputs of the antecedent agents $\{O_m | m \in A(n)\}$. Please refer to \S\ref{sec:multi_agent_details} in Appendix for more details regarding the concrete role-based multi-agent system architecture we adopt in our work.

Consider a training set $D_{train} = \{X_i\}_{i=1}^{|\mathcal{T}|}$ without ground-truth labels drawn from a software development task $\mathcal{T}$, the multi-agent system $G$ takes each $X_i$ as input, processes $X_i$ given the graph structure and current agents prompt group, and outputs a final response $Y_i = G(X_i, \mathcal{P})$. Given a user-defined evaluation dimension based on the user's need, a critic mechanism $f$ targeting this evaluation dimension generates a scalar score $U_i$ that directly assesses the utility of $Y_i$ (performance of the multi-agent system) and natural language feedback $I_i$ explaining the scalar score based on the task input: $U_i, I_i = f(Y_i, X_i)$. In our work, we study 5 evaluation dimensions as introduced in \S\ref{sec:optimization_dimensions}, and we will describe the concrete implementations of their corresponding critic mechanisms in \S\ref{sec:eval_implementation}. Our goal is to find an optimal prompt group $\mathcal{P}^*$ drawn from the natural language space such that the expectation of utility $U_i$ is maximized over $D_{train}$ utilizing $I_i$:
\[\mathcal{P}^* = \argmax_\mathcal{P} \E_{x_i \sim D_{train}}[U_i]\]
Finally, we use the optimized agent prompt group $\mathcal{P}^*$ to perform evaluation on a testing set based on utility scores generated by the same critic mechanism: $\E_{x_i \sim D_{test}}[U_i]$.

\subsection{Optimization Pipeline}
\label{sec:optimization_pipeline}
\begin{figure}[!t]
\begin{center}
    \includegraphics[scale=0.6]{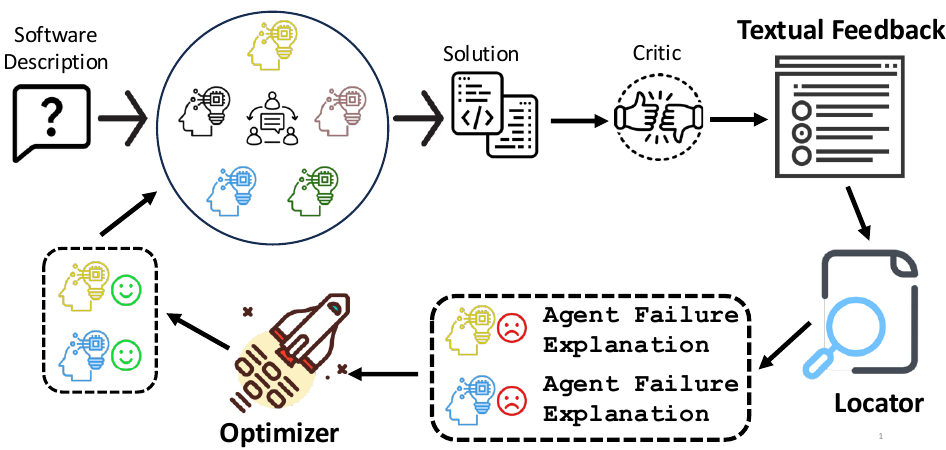}
    \caption{A high-level workflow of our two-step optimization pipeline. Please refer to \S\ref{sec:optimization_pipeline} for more details.}
    % A locator first locates the underperforming agents and generates explanations of their failures utilizing textual feedback. Then, an optimizer optimizes the system prompts of the underperforming agents using their failure explanations. 
    \label{fig:pipeline}
\end{center}
\end{figure}

We propose a two-step pipeline to optimize a role-based multi-agent system. As shown in Figure \ref{fig:pipeline}, utilizing natural language feedback $I$ \img{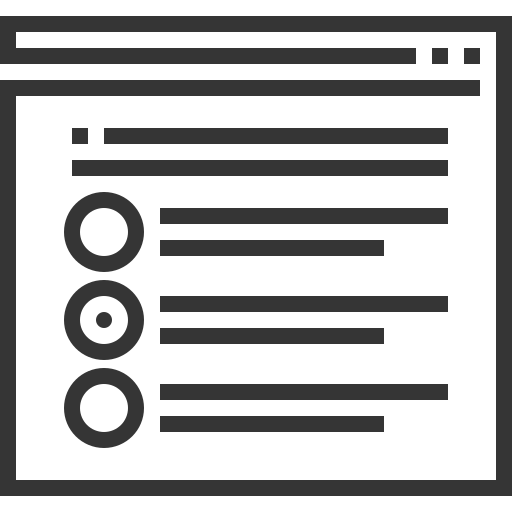} of the multi-agent system's output \img{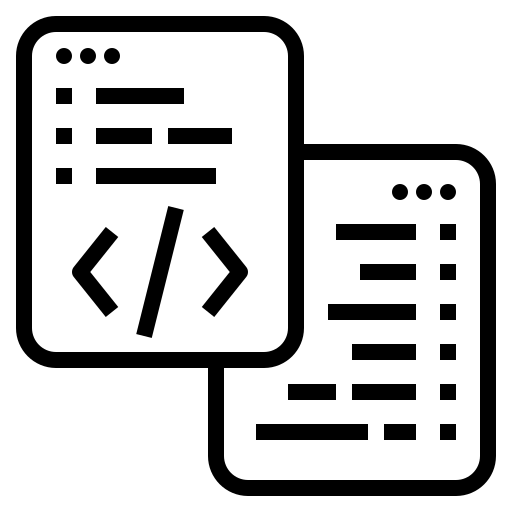} generated by the critic mechanism $f$ \img{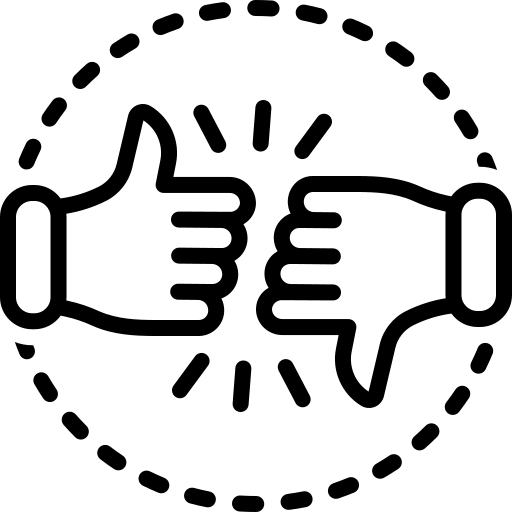}, we perform prompt optimization for participation agents in the system. Specifically, we first use a locator $\mathcal{L}$ \img{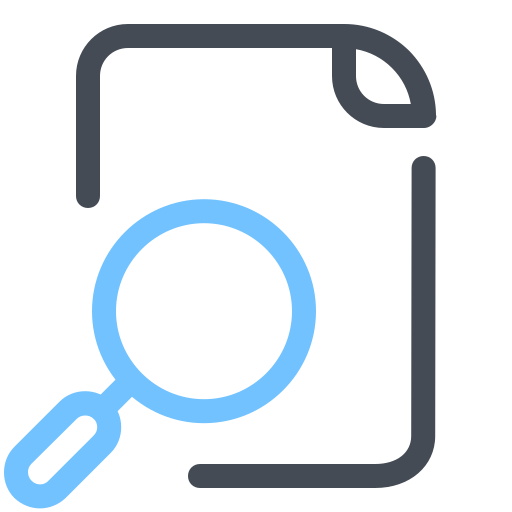} to identify the underperforming agents, \img{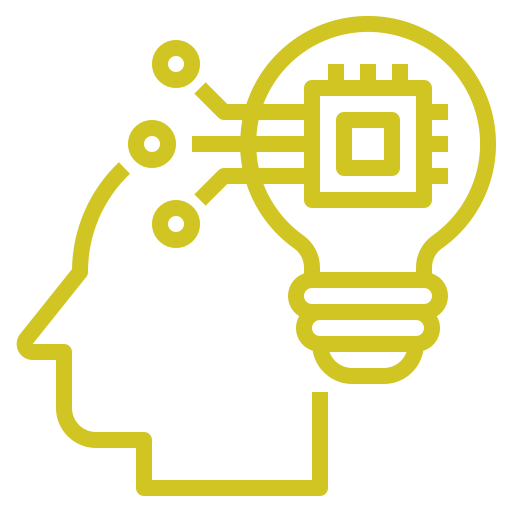} and \img{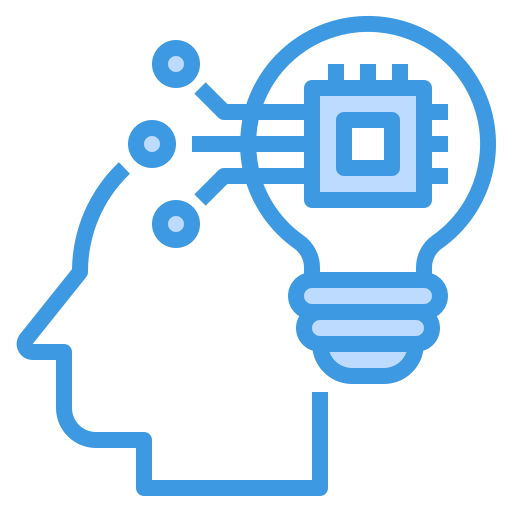}, that are not performing well and generate explanations of their failures, utilizing textual feedback $I$ (\S\ref{sec:locator_step}). Then, an optimizer $\mathcal{O}$ \img{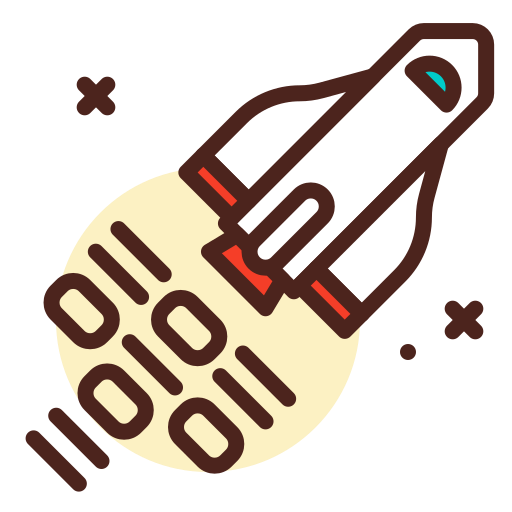} optimizes the system prompts of the identified underperforming agents utilizing explanations of their failures (\S\ref{sec:optimize_step}).

\subsubsection{Locating}
\label{sec:locator_step}
Natural language feedback $I$ for the multi-agent system's output provides guidance regarding which directions to improve agents in the system. However, it is hard to improve individual agents solely based on global feedback of the final collaboration output. As a result, we design an LLM-based locator $L$ \img{figures/locator.png} capable of generating fine-grained explanations to guide agent optimization. Our locator takes three major components as inputs: software task description $X$, global natural language feedback $I$, and multi-agent collaboration details $G$. Collaboration details include two major components: role descriptions and communication trajectories of all participation agents.
With a carefully designed prompt (see Figure \ref{fig:locator_prompt} in Appendix), the locator focuses on the negative aspects of the global feedback and navigates to underperforming agents $N_f \subseteq N$ responsible for the negative aspects of feedback. 
The locator also generates specific failure explanations $E = \{E_n | n \in N_f\}$ of identified underperforming agents as fine-grained signals to better guide the later optimization step. 
Overall, the input and output workflow of the locator is $N_f, E = L(X, I, G)$.

\subsubsection{Optimizing}
\label{sec:optimize_step}
After identifying the underperforming agents and fine-grained explanations of their failures from the locator in the previous step, we utilize an LLM-based optimizer $O$ \img{figures/optimizer.png} to optimize the system prompts of underperforming agents. For each underperforming agent $n$, the optimizer takes two critical components as input: input-output pair $M_n$ and fine-grained failure explanations $E_n$ of this agent. The input-output pair includes the system message $P_n$, all user messages, and the output of the agent. We carefully design the prompt (see Figure \ref{fig:optimizer_prompt} in Appendix) to guide the optimizer $O$ to optimize the system prompt of underperforming agents in the direction where the fine-grained failure explanations can be mitigated: $P_n = O(P_n, E_n, ...)$.

\subsection{Optimization Settings}
\label{sec:optimization_setting}

Based on our proposed two-step optimization pipeline, we proceed to investigate the impact of various optimization settings on the performance of the multi-agent system using two comparison groups: online against offline (\S\ref{sec:offline_advantage}) and individual against group (\S\ref{sec:completeness}) optimization. For group optimization setting, we further investigate two different prompting methods: one-pass against multi-pass optimization prompting.

\subsubsection{Online against Offline Optimization}
\label{sec:offline_advantage}
Inspired by the main difference between online and offline reinforcement learning \cite{levine2020offline} where data for learning is collected in real-time by agents interacting with the environment under online setting, whereas collected beforehand under offline setting, we apply the online versus offline setting to our study. The difference between our online and offline settings lies in how the textual feedback is collected during each optimization step (each training instance in our case). For the online setting, we use the optimized prompts at the current step to derive the solution code so that agents must interact with the environment (the critic) to retrieve real-time feedback. However, for offline setting, we use the default initial agent prompts to derive solution codes and retrieve feedback for all training instances beforehand. This offline feedback collection process fits into open-ended tasks like software development, where high-quality human-annotated feedback can be collected beforehand.

\subsubsection{Individual against Group Optimization}
\label{sec:completeness}
During the locating step, multiple agents are usually identified as underperforming agents. We aim to investigate whether optimizing all of them or disentangling the optimization by individually updating each agent is more effective. One can image group optimization as a complete optimization process as it optimizes all components in each step; however, it also potentially brings problems such as overfitting, and individual optimization setting in our study can mitigate this concern by reducing the optimization complexity and gradually optimizing one agent at a time. Concretely, we randomly sample one underperforming agent for optimization and leave other underperforming agents untouched during each optimization step.

\noindent \textbf{One-Pass versus Multi-Pass Prompting.} \hspace{4pt}
For group optimization, we study two prompting methods. For all identified underperforming agents, we optimize each with a separate LLM inference call, which we call multi-pass prompting. We can also optimize all agents jointly with a single LLM inference call, which we call one-pass prompting. Multi-pass prompting could be more accurate than one-pass prompting, as one-pass prompting introduces irrelevant information about other agents. However, one-pass prompting could utilize the interconnection between agent components, and it is more efficient as it consumes fewer API calls.

\section{Experiments}

\subsection{Dataset}
We use SRDD (Software Requirement Description Dataset) \cite{qian-etal-2024-chatdev} as the software requirement descriptions dataset in our study. SRDD comprises 1,200 software task prompts extracted from ChatGPT, spanning 5 major categories, including education, work, life, game, and creation. We randomly shuffle and split the entire dataset into train, development, and test splits with a ratio of 6:2:2. 

\subsection{Critic Mechanism Implementation}
\label{sec:eval_implementation}
We explore various evaluation dimensions pertinent to software development tasks. For each dimension, a critic mechanism generates both a scalar utility score and natural language feedback to the solution as described in \S\ref{sec:formulation}. During implementation, scores and feedback are generated simultaneously or separately for different dimensions. Overall, they are generated using two high-level methodologies. First is rule-based method, where scores or feedback are generated based on heuristic rules or external tools. Second is model-based method, following recent works of LLMs as judges \cite{zheng2023judging, dubois2023alpacafarm, mcaleese2024llm}, where we guide LLMs to evaluate solution code with carefully designed prompts.  

For functionality, robustness, and test case coverage dimension, we generate scores on a scale from 0 to 10 and natural language feedback at the same time using GPT-4 \cite{Achiam2023GPT4TR} as judge. For the documentation dimension, we use GPT-4 to generate natural language feedback; however, we directly use the number of lines of comments and docstrings in the solution code as the scalar score. 
For code style violation dimension, we utilize an external tool, \texttt{pycodestyle}\footnote{\url{https://pycodestyle.pycqa.org/en/latest/}}, to check the software code against style conventions in PEP 8. 
% For each style violation, the checker outputs the line number where the violation occurs with its explanation in natural language format, for example, \textit{multiple spaces before operator}. 
We define the total number of violations and the corresponding explanations the checker identifies as score and feedback.
Please refer to Appendix \ref{sec:eval_prompt} for more details of prompts for obtaining scores and feedback using GPT-4.

\subsection{Experiment Setting}
We use \texttt{gpt-4-0613} version of GPT-4 as the LLM everywhere in our study with a temperature of 0.1.
We randomly sample 5 task descriptions \footnote{We found more steps of optimization unnecessary; please refer to \S\ref{sec:more_steps} in Appendix for more details} from the training set. At each optimization step, we optimize the current agent system prompts to a new group of prompts, which will be optimized in the next step.
Unless mentioned explicitly, we always randomly sample 100 task descriptions from the testing set to report evaluation results due to budget constraints. 
% For the role-based multi-agent system to optimize, we adopt the vertical decision-making structure \cite{chen2024agentverse} where a solver agent is responsible for writing the solution code and other reviewer agents are responsible for providing feedback to the solver agent for later revision.
% Please refer to Appendix \ref{sec:multi_agent_details} for more details of the role-based multi-agent system.

\begin{table}[!t]
\centering
\scalebox{0.73}{
    \begin{tabular}{@{}lccccc@{}}
    \toprule
     & \textbf{Functionality ($\uparrow$)} & \textbf{Robustness ($\uparrow$)} & \textbf{Coverage ($\uparrow$)} & \textbf{Documentation  ($\uparrow$)} & \textbf{Violation  ($\downarrow$)} \\ \midrule
    Unoptimized & 6.90 & 6.75 & 0.32 & 3.80 & 6.62 \\
    One-Shot & 6.66 & 7.47 & 7.00 & 15.33 & 3.80 \\
    Direct Optimization (TextGrad) & 6.74 & 7.11 & 6.31 & 16.07 & 6.90 \\
    \midrule
    \multicolumn{6}{c}{\cellcolor[HTML]{EFEFEF}\textit{Our Optimization}} \\
    \midrule
    \textit{Offline Setting} & & & & & \\
    \hspace{3mm} Individual & 6.81 & 7.20 & 7.27 & 19.46 & 5.18 \\
    \hspace{3mm} Group w/ Multi-Pass & {\ul 7.22} & 7.63 & {\ul 7.64} & 17.92 &4.35  \\
    \hspace{3mm} Group w/ One-Pass & 7.06 & \textbf{7.77} & 6.99 & 19.92 & 4.82 \\
    \textit{Online Setting} & & & & & \\
    \hspace{3mm} Individual & 7.02 & 7.55 & 6.95 & 20.14 & 5.34 \\
    \hspace{3mm} Group w/ Multi-Pass & \textbf{7.26} & 7.60 & 6.48 & {\ul 20.15} & \textbf{3.03} \\
    \hspace{3mm} Group w/ One-Pass & \textbf{7.26} & {\ul 7.74} & \textbf{7.81} & \textbf{20.33} & {\ul 3.24} \\
    
     \bottomrule
    \end{tabular}
}
\caption{Evaluation scores under all optimization settings across all evaluation dimensions for baselines and our proposed two-step optimization pipeline. Bold numbers indicate best-performing results, and underlined numbers indicate second-best results. Note that the score range for the first three dimensions is 0-10, and the last two are simply positive integers.}
\label{tab:main_table}
\end{table}

\subsection{Baselines}
We consider the unoptimized system and two baselines for comparisons. \textbf{Unoptimized}: we use default agent system prompts to run the pipeline for code generation without optimization directly. \textbf{One-shot}: We randomly sample one agent communication trajectory and feedback pair from the training set into the system prompts of all agents as a demonstration and ask them to avoid making similar mistakes presented in the feedback. \textbf{Direct optimization (TextGrad)}: We consider another baseline that directly optimizes all system prompts given textual feedback. This aligns with TextGrad \cite{Yuksekgonul2024TextGradA}, which backpropagates textual feedback to improve individual components of a compound AI system. Note that the key difference lies in that TextGrad does not identify the underperforming agents as the locator in our framework does and directly utilizes the textual feedback for optimization.
Another related work is DSPy \cite{khattab2024dspy}, a programming model that abstracts LLM pipelines as text transformation graphs, i.e., imperative computational graphs where LLMs are invoked through declarative modules. However, the evaluation metric must output numerical values instead of textual feedback, which cannot fit our setting. Therefore, we leave adapting it to our setting as future work.
% We first compare our approach with the unoptimized system using default agent system prompts to show the effectiveness of our pipeline.
% Given that there is no previous work studying group optimization for role-based multi-agent systems targeting software development tasks, we propose two optimization baselines under our setting. The first is a one-shot optimization baseline where we directly incorporate one randomly sampled agent communication trajectory and feedback pair from the training set into the system prompts of all agents as a demonstration and ask them to avoid making similar mistakes presented in the feedback. The second is an implicit reflection-based prompt optimization method. Inspired by OPRO \cite{yang2024large}, we prompt an LLM-based optimizer to implicitly reflect on the textual feedback and communication trajectories to directly refine current agent prompts during each optimization step with a single inference call. We name it as implicit reflection.
% Due to the context length limit of GPT-4, we can only study the one-shot setting instead of the few-shot setting.

\begin{figure}[!t]
    \centering
    % \captionsetup{skip=0pt}
    \begin{tikzpicture}[scale=0.70]

        \begin{axis}[
            name=functionality_plot,
            title={\normalsize Functionality ($\uparrow$)},
            ylabel={\small Evaluation Scores},
            xlabel={\small Optimization Steps (\%)},
            xmin=-5, xmax=105,
            ymin=5.9, ymax=8.6,
            xtick={0,20,40,60,80,100},
            xticklabels={0\%,20\%,40\%,60\%,80\%,100\%},
            ytick distance=0.5,
            xmajorgrids=true, ymajorgrids=true,            
            height=100pt,
            width=140pt,
            every axis/.append style={font=\tiny},
            % every axis plot/.append style={thick},
            legend style={
                at={(0.97,0.03)},
                anchor=south east,
                font=\scriptsize,
                draw=none,
                fill=none,
                legend columns=-1,
                /tikz/every even column/.append style={column sep=0.2cm}
            },
        ]
            \addplot[color=BrickRed,mark=triangle*,]  
            coordinates {(0,6.9)(20,7.6)(40,7.4)(60,7.6)(80,7)(100,8)};
            \addplot[color=Violet,mark=*,]  
            coordinates {(0,6.48)(20,6.93)(40,7.13)(60,7.27)(80,6.8)(100,7.53)};
            \legend{train, dev}
        \end{axis}
    
        \begin{axis}[
            name=robustness_plot,
            at={($(functionality_plot.north east)+(15,0pt)$)}, anchor=north west,
            title={\normalsize Robustness ($\uparrow$)},
            xlabel={\small Optimization Steps (\%)},
            xmin=-5, xmax=105,
            ymin=6.4, ymax=8.6,
            xtick={0,20,40,60,80,100},
            xticklabels={0\%,20\%,40\%,60\%,80\%,100\%},
            ytick distance=0.5,
            xmajorgrids=true, ymajorgrids=true,            
            height=100pt,
            width=140pt,
            every axis/.append style={font=\tiny},
            % every axis plot/.append style={thick},
            legend style={
                at={(0.97,0.97)},
                anchor=north east,
                font=\scriptsize,
                draw=none,
                fill=none,
                legend columns=-1,
                /tikz/every even column/.append style={column sep=0.2cm}
            },
        ]
            \addplot[color=BrickRed,mark=triangle*,]  
            coordinates {(0,7.2)(20,7)(40,6.8)(60,8)(80,7.8)(100,7.4)};
            \addplot[color=Violet,mark=*,]  
            coordinates {(0,7.1)(20,7.2)(40,7.53)(60,7.6)(80,7.63)(100,7.67)};
            \legend{train, dev}
        \end{axis}

        \begin{axis}[
            name=coverage_plot,
            at={($(robustness_plot.north east)+(15,0pt)$)}, anchor=north west,
            title={\normalsize Test Case Coverage ($\uparrow$)},
            xlabel={\small Optimization Steps (\%)},
            xmin=-5, xmax=105,
            ymin=-0.4, ymax=10.4,
            xtick={0,20,40,60,80,100},
            xticklabels={0\%,20\%,40\%,60\%,80\%,100\%},
            ytick distance=2,
            xmajorgrids=true, ymajorgrids=true,            
            height=100pt,
            width=140pt,
            every axis/.append style={font=\tiny},
            % every axis plot/.append style={thick},
            legend style={
                at={(0.97,0.03)},
                anchor=south east,
                font=\scriptsize,
                draw=none,
                fill=none,
                legend columns=-1,
                /tikz/every even column/.append style={column sep=0.2cm}
            },
        ]
            \addplot[color=BrickRed,mark=triangle*,]  
            coordinates {(0,0.2)(20,4)(40,6.8)(60,6.2)(80,7.2)(100,7.8)};
            \addplot[color=Violet,mark=*,]  
            coordinates {(0,0.4)(20,4.6)(40,7.07)(60,5.53)(80,6.3)(100,7.23)};
            \legend{train, dev}
        \end{axis}

        \begin{axis}[
            name=documentation_plot,
            at={($(coverage_plot.north east)+(15,0pt)$)}, anchor=north west,
            title={\normalsize Documentation ($\uparrow$)},
            xlabel={\small Optimization Steps (\%)},
            xmin=-5, xmax=105,
            ymin=-1, ymax=31,
            xtick={0,20,40,60,80,100},
            xticklabels={0\%,20\%,40\%,60\%,80\%,100\%},
            ytick distance=5,
            xmajorgrids=true, ymajorgrids=true,            
            height=100pt,
            width=140pt,
            every axis/.append style={font=\tiny},
            % every axis plot/.append style={thick},
            legend style={
                at={(0.97,0.03)},
                anchor=south east,
                font=\scriptsize,
                draw=none,
                fill=none,
                legend columns=-1,
                /tikz/every even column/.append style={column sep=0.2cm}
            },
        ]
            \addplot[color=BrickRed,mark=triangle*,]  
            coordinates {(0,2)(20,11.8)(40,26)(60,23.6)(80,23.6)(100,21)};
            \addplot[color=Violet,mark=*,]  
            coordinates {(0,5.9)(20,14.13)(40,19.43)(60,18.33)(80,20.9)(100,20.8)};
            \legend{train, dev}
        \end{axis}

        \begin{axis}[
            name=compliance_plot,
            at={($(documentation_plot.north east)+(15,0pt)$)}, anchor=north west,
            title={\normalsize Violation ($\downarrow$)},
            xlabel={\small Optimization Steps (\%)},
            xmin=-5, xmax=105,
            ymin=-0.4, ymax=10.4,
            xtick={0,20,40,60,80,100},
            xticklabels={0\%,20\%,40\%,60\%,80\%,100\%},
            ytick distance=2,
            xmajorgrids=true, ymajorgrids=true,            
            height=100pt,
            width=140pt,
            every axis/.append style={font=\tiny},
            % every axis plot/.append style={thick},
            legend style={
                at={(0.97,0.03)},
                anchor=south east,
                font=\scriptsize,
                draw=none,
                fill=none,
                legend columns=-1,
                /tikz/every even column/.append style={column sep=0.2cm}
            },
        ]
            \addplot[color=BrickRed,mark=triangle*,]  
            coordinates {(0,8.6)(20,5.8)(40,5)(60,4.8)(80,6.4)(100,2.6)};
            \addplot[color=Violet,mark=*,]  
            coordinates {(0,7)(20,7.07)(40,4.13)(60,4.57)(80,3.43)(100,4.27)};
            \legend{train, dev}
        \end{axis}

    \end{tikzpicture}
    \caption{Training and development evaluation score curves analysis for all evaluation dimensions under online group optimization with one-pass prompting.}
    \label{fig:optim_curve_all_dims}
\end{figure}

\subsection{Main Results}
We show the main results in Table \ref{tab:main_table}.
Both one-shot and direct optimization baselines cannot consistently outperform unoptimized system across all evaluation dimensions. They both fail on the functionality dimension, which is probably the most important evaluation dimension for software development. Direct optimization also fails on the code violation dimension.
However, our proposed method effectively optimizes the multi-agent system evaluated on all optimization dimensions and settings except only for offline individual setting on the functionality dimension, where we observe only 0.1 behind the unoptimized system. Our best-performing optimization setting, online group optimization with one-pass prompting, is always better than the two baselines.
Our method outperforms the one-shot baseline in 22 out of the total of 30 cases across all evaluation dimensions and optimization settings, and it always outperforms the direct optimization baseline.

Next, we compare the performance of our optimization pipeline under all optimization settings. 
First, the most effective strategy seems to be online group optimization with one-pass prompting, whose performance is consistently among the top two across all evaluation dimensions. 
Secondly, individual optimization is definitely worse than group optimization in most cases; however, it is still an effective optimization strategy compared with unoptimized system in all cases only except for offline setting for functionality dimension. This means optimizing all underperforming agents at each step is a better practice than gradually optimizing a single agent component along the optimization steps under our case study. It is possible because an LLM-based multi-agent system is still not complex enough to bring up issues like overfitting compared with more complicated systems such as neural networks.
Thirdly, although counterintuitive in our case study, the offline setting is still an effective optimization strategy. Offline setting is generally worse than the online setting only across 2 out of 5 evaluation dimensions. As discussed in Section \ref{sec:offline_advantage}, this enables human intervention by providing high-quality feedback annotation beforehand. This is even more beneficial when the required training data size is only a few, meaning less human-effort required for annotation. This is very similar to our case, and we leave investigating whether human-annotated high-quality feedback leads to even better optimization for future work. 
Finally, we don’t observe a consistent performance difference between one-pass and multi-pass optimization prompting for group optimization setting across all evaluation dimensions. This probably suggests that although the single LLM call of the one-pass prompting contains irrelevant information about other agents, it does not affect the overall optimization process under our current setup. As a result, one could choose one-pass optimization for higher efficiency without sacrificing performance.

\subsection{Analyses}

\subsubsection{Optimization Curve Analysis}

\begin{wrapfigure}{R}{0.3\textwidth}
    \centering
    \begin{tikzpicture}[scale=0.6]
        \begin{axis}[
            name=train_plot,
            xlabel={\normalsize Optimization Steps}, 
            ylabel={\normalsize Evaluation Scores ($\downarrow$)},
            xmin=-2, xmax=102,
            ymin=1.6, ymax=12.5,
            xtick={0,20,40,60,80,100},
            xticklabels={0\%,20\%,40\%,60\%,80\%,100\%},
            ytick distance=2,
            xmajorgrids=true, ymajorgrids=true,            
            height=140pt,
            width=210pt,
            every axis plot/.append style={thick},
            every axis/.append style={
                font=\small,
                % line width=1pt,
                % tick style={line width=3pt}
            },
            legend style={
                font=\small,
                at={(0.5,1.05)},
                anchor=south,
                legend columns=2,
                % column sep=0.5cm,
                % /tikz/every even column/.append style={column sep=0.5cm},
                nodes={scale=0.75},
                legend cell align={left},
            },
        ] 
            \addplot[color=OliveGreen,mark=*,]  
            coordinates {(0,8.6)(20,6)(40,5.6)(60,3.8)(80,2.6)(100,2.4)};
            \addplot[color=Bittersweet,mark=square*,]  
            coordinates {(0,8.6)(20,5.8)(40,5)(60,4.8)(80,6.4)(100,2.6)};
            \addplot[color=Brown,mark=triangle*,]  
            coordinates {(0,8.6)(20,5.6)(40,3.4)(60,8)(80,6.2)(100,4.2)};
            \addplot[color=NavyBlue,mark=diamond*,]  
            coordinates {(0,8.6)(20,7.4)(40,11)(60,10)(80,4.8)(100,5)};
            \legend{Online Multi-Pass, Online One-Pass, Offline Multi-Pass, Offline One-Pass}
        \end{axis}
    \end{tikzpicture}
    \caption{Training curve analysis for code style violation evaluation dimension under major optimization settings.}
    \label{fig:optim_curve_all_settings}
\end{wrapfigure}

In this section, we plot the optimization curves of the average evaluation scores on the training and development set with respect to each optimization step. Due to budget limits, we are not able to plot curves for all development examples, so we randomly sample 30 examples from the development set. 
Figure \ref{fig:optim_curve_all_dims} shows the optimization curves for all evaluation dimensions under the online group optimization with one-pass prompting setting. We chose this setting as it gives superior performances compared with other settings as discussed in previous section.
First, we can tell that there is no over-fitting happening during training, as the trends for both training and development curves are consistent across all evaluation dimensions. Second, we observe that complete model-based critic mechanisms (first three) tend to show less stable training than rule-based critic mechanisms (last two), as they oscillate more often. Finally, it shows that training curves are less stable than development curves, possibly due to the sparsity of training data compared with development data.
Figure \ref{fig:optim_curve_all_settings} shows the optimization training curves under major optimization settings (we use group optimization as default) for code style violation dimension since it has a complete model-free score and feedback generation process to avoid potential bias raised by model-based evaluation. We observe that online setting shows much better stability than offline settings under both one-pass and multi-pass prompting, as their curves oscillate much less.

\subsubsection{Starting from "Empty" Prompts}

\begin{wrapfigure}{R}{0.35\textwidth}
    \centering
    \scalebox{0.70}{
        \begin{tabular}{@{}cccc@{}}
        \toprule
         & \textbf{\small Unoptimized} & \textbf{\small Default} & \textbf{\small Empty} \\ \midrule
        Func. ($\uparrow$) & 6.90 & 7.26 & 7.06 \\
        Rob. ($\uparrow$) & 6.75 & 7.74 & 7.77 \\
        Cov. ($\uparrow$) & 0.32 & 7.81 & 7.31 \\
        Doc. ($\uparrow$) & 3.80 & 20.33 & 19.8 \\
        Comp. ($\downarrow$) & 6.62 & 3.24 & 4.61 \\ \bottomrule
        \end{tabular}
    }
    \caption{The effect of starting optimization from an "empty" agent prompt instead of a default prompt.}
    \label{tab:start_from_bad}
\end{wrapfigure}

Instead of starting optimization from a default prompt, we study whether our optimization pipeline is still effective when the prompt to optimize starts from empty. Note that the "empty" prompts are not completely empty as they still contain very basic contexts (prompts in black) as shown in Figures \ref{fig:solver_prompt} and \ref{fig:reviewer_prompt}. We slightly increase the optimization steps from the default number of 5 to 8. 
We analyze online group optimization with one-pass promoting across all evaluation dimensions. As shown in Table \ref{tab:start_from_bad}, starting from an "empty" prompt, our pipeline is still able to optimize the multi-agent system to a level that is on par or just slightly worse than the system optimized starting from an informative default starting prompt. 
%  We don't deliberately choose this 8 or say we choose it based on eval set

\subsubsection{Case Study}

We provide optimized agent prompts at the final step for functionality, robustness, and code style violation evaluation dimensions for the two agent roles: the programmer agent in Table \ref{tab:programmer_agent_case} and the software test engineer agent in Table \ref{tab:test_engineer_agent_case} in the Appendix. We generally observe agent prompts being optimized under the desired evaluation dimension, as highlighted in green text. 
For example, under the functionality dimension, the system prompt of the solver agent is optimized to contain "\textit{ensure that all functionalities mentioned in the task description are implemented}", and system prompt of the reviewer agent is optimized to contain "\textit{provide more detailed feedback on any missing functionalities or areas for improvement}". 
Under the dimension of code violation, the system prompt of the solver agent is optimized to contain "\textit{break down long lines of code into multiple lines to improve readability and maintainability}", and the reviewer agent is optimized to contain "\textit{review the code for adherence to style guides like PEP8, including line length, and provide feedback on this aspect}". In future work, we aim to improve the generalization of the optimized prompt. For example, we observe that the optimized prompt sometimes contains instance-specific content of the training instances such as "\textit{consider the lifecycle of any temporary files created during the
process}".

% In future work, 
% However, we also notice a current problem: as shown in the red text, the optimized prompts might contain instance-specific content that does not apply to general software development tasks even though we deliberately prompt the optimizer to think generally instead of focusing on the current task only. We leave mitigating this problem to future work.

% \begin{table*}[]
% \centering

% \begin{subtable}{1\textwidth}
% \centering
% \scalebox{1}{
%     \begin{tabular}{@{}lcccccc@{}}
%     \toprule
%     \multirow{2}{*}{} & \multicolumn{2}{c}{Functionality} & \multicolumn{2}{c}{Robustness} & \multicolumn{2}{c}{Test Case Coverage} \\ \cmidrule(l){2-3} \cmidrule(l){4-5} \cmidrule(l){6-7} 
%      & Unoptimized & Optimized & Unoptimized & Optimized & Unoptimized & Optimized \\ \midrule
%     Default Prompt & 6.90 & 7.26 & 6.75 & 7.74 & 0.32 & 7.81 \\
%     Empty Prompt & 6.18 & 7.06 & 6.92 & 7.77 & 0.30 & 7.31 \\ \bottomrule
%     \end{tabular}
% }
% % \caption{here goes caption}
% \end{subtable}

% \bigskip

% \begin{subtable}{1\textwidth}
% \centering
% \scalebox{1}{
%     \begin{tabular}{@{}lcccc@{}}
%     \toprule
%     \multirow{2}{*}{} & \multicolumn{2}{c}{PEP 8} & \multicolumn{2}{c}{Comment} \\ \cmidrule(l){2-3} \cmidrule(l){4-5} 
%      & Unoptimized & Optimized & Unoptimized & Optimized \\ \midrule
%     Default Prompt & 6.62 & 3.24 & 3.80 & 20.33 \\
%     Empty Prompt & 6.20 & 4.61 & 4.59 & 19.8 \\ \bottomrule
%     \end{tabular}
% }
% % \caption{here goes caption}
% \end{subtable}

% \end{table*}

\begin{table*}[!t]
    \centering
    \scalebox{0.95}{
        \begin{tabular}{p{1\linewidth}}
             \toprule
             \textbf{Functionality} \\
             \midrule
              As a Programmer, your task is to provide a comprehensive solution to the given software task. Your solution should be versatile, capable of handling different sports, player positions, and strategies. It should also allow users to drag and drop players to specific positions and add notes and annotations to each play. Ensure that all methods outlined in the initial structure are fully implemented and functional. Pay special attention to the user interface and ensure it is user-friendly. {\color{ForestGreen} \textbf{Test your code to ensure it works as expected and meets all the requirements of the task description. Additionally, ensure that all functionalities mentioned in the task description are implemented, and consider the user experience when designing the interface and functionality of the software.}} After receiving feedback from other agents, make sure to incorporate their suggestions into your final solution. \\
             \midrule
             \textbf{Robustness} \\
             \midrule
             As a Programmer, your task is to provide a new solution code to the given software task. If the history is not empty, your new solution code must be based upon your previous solution and teammates' feedback in the history. {\color{ForestGreen} \textbf{While developing your solution, ensure to incorporate robust error handling, especially for user inputs in any user interface elements. Consider all possible edge cases and potential user errors to enhance the robustness of your solution. Additionally, pay attention to the feedback from your teammates regarding the robustness of your code, particularly in areas such as error handling, input validation, and handling of unexpected exceptions.}} Also, consider the lifecycle of any temporary files created during the process and ensure they are properly managed to prevent unnecessary storage usage. {\color{ForestGreen} \textbf{Remember to think thoroughly about the different ways the software could be used or misused, and add appropriate error handling and input validation to cover these scenarios.}} Furthermore, consider the integrity and validity of the data being processed, including checks for invalid or duplicate inputs, and robust handling of data storage and retrieval. \\
             \midrule
             \textbf{Violation} \\
             \midrule
             As a programmer, your task is to provide a new solution code to the given software task. If the history is not empty, your new solution code must be based upon your previous solution and teammates' feedback in the history. {\color{ForestGreen} \textbf{While writing the code, ensure that it adheres to the PEP8 style guide, especially the rule about maximum line length. Break down long lines of code, including comments and docstrings, into multiple lines to improve readability and maintainability. Use a linter or code formatter to automatically check and correct your code style. Also, manually check your code against the PEP8 style guide before submitting it.}} In addition, when revising your code based on feedback, make sure to address all the points raised by your teammates, especially those related to code style and organization. \\
             \bottomrule 
        \end{tabular}
    }
    \caption{We list optimized system prompts for the solver agent whose role is described as a programmer along functionality, robustness, and code violation evaluation dimensions. Green text indicates agent prompts are being optimized under the evaluation dimension. 
}
    \label{tab:programmer_agent_case}
\end{table*}
\section{Conclusion}

In this work, we present a case study on group behavior optimization with multiple LLM-based agents utilizing natural language feedback on software development. We first propose a two-step optimization framework to effectively optimize a role-based multi-agent system under various user-defined evaluation dimensions. We then investigate the impacts of various optimization settings and provide valuable insights regarding group optimization behaviors under those settings.

% \section*{Acknowledgments}

% \section*{Ethics Statement}

\bibliography{colm2025_conference}
\bibliographystyle{colm2025_conference}

\appendix

\section{Role-based Multi-Agent System}
\label{sec:multi_agent_details}
Various decision-making structures among agents \cite{chan2024chateval,wu2024autogen} have been investigated.
Following AgentVerse\cite{chen2024agentverse}, we adopt the vertical decision-making structure as it is a better fit for software development. Inside the vertical structure, given the software task description $X$, a solver agent $S$ proposes a solution $Y_t$ at iteration $t$. Other agents, as reviewer agents $\mathcal{R}$, provide feedback $\mathcal{F} = \{r_t^i = R_i(Y_t) | R_i \in \mathcal{R}\}$ regarding solution $Y_t$ to the solver agent. Finally, the solver agent refines its solution based on the feedback $Y_{t+1} = S(X, Y_t, \mathcal{F})$. Such a review iteration can go on for a few rounds. In the current case study, we use a total of 2 reviewer agents and a single review iteration.
To determine the concrete role descriptions for solver and reviewer agents, we utilize a recruiter agent $A$ to select role descriptions for the solver and reviewer agents from a pre-defined expert pool based on the current task description. The agent role description pool we use is adapted from ChatDev \cite{qian-etal-2024-chatdev}. 
% Example role descriptions include "\textit{Programmer who can can write/create computer software with ...}", "\textit{Chief product officer who is responsible for all product-related matters including ...}", and "\textit{Code reviewer who can help programmers to assess source codes for ...}"
In our study, we aim to optimize the system prompts of all agents in the vertical decision-making structure.

\begin{figure}
    \centering
    \begin{tcolorbox}[left=5pt,right=5pt,colback=white,colframe=black,boxrule=1pt,fontupper=\ttfamily]
        \textbf{System Prompt}: \\
        You are \$\{role description\}. You are in a multi-agent collaboration environment. \\

        You are given a software development task description: \\
        \$\{task description\} \\ 
        
        You will also be given the chat history of you and other teammates (history could be empty). \\ 

        {\color{red} <TO IMPROVE> Your task is to provide a new solution code to the given software task. If the history is not empty, your new solution code must be based on your previous solution and teammates' feedback in the history. </TO IMPROVE>} \\ 

        \textbf{User Prompt}: \\
        Start your response now and write your code step by step in \texttt{\textasciigrave \textasciigrave \textasciigrave}python markdown quotes.
    \end{tcolorbox}
\caption{System and user prompt for solver agent whose responsibility is to write the main solution code given a software task description.}
\label{fig:solver_prompt}
\end{figure}

\begin{figure}
    \centering
    \begin{tcolorbox}[left=5pt,right=5pt,colback=white,colframe=black,boxrule=1pt,fontupper=\ttfamily]
        \textbf{System Prompt}: \\
        You are \$\{role description\}. You are in a multi-agent collaboration environment aiming to solve a given software development task: \$\{task description\}. \\
        
        You are also given the chat history of you and other teammates below. \\

        {\color{red} <TO IMPROVE> Based only on your expertise, please provide your feedback on the most recent solution code to the software task given. Ensure your feedback is specific and detailed enough instead of just general opinions. </TO IMPROVE>} \\ 

        \textbf{User Prompt}: \\
        Start your response now.

    \end{tcolorbox}
\caption{System and user prompt for reviewer agent whose responsibility is to review the solution code written by solver agent and provide feedback to solver agent given a software task description.}
\label{fig:reviewer_prompt}
\end{figure}

The concrete prompt for the solver agent is shown in Figure \ref{fig:solver_prompt}, and the prompt for reviewer agents is shown in Figure \ref{fig:reviewer_prompt}. The system prompts we aim to optimize are shown in red, and we wrap them with \texttt{<TO IMPROVE>} tags to instruct LLM to optimize only text between the tags so that other text is untouched. For the recruiting stage, the role description pool we use is directly presented in the prompt for the recruiter agent, as shown in Figure \ref{fig:recruiter_prompt}. Note that we restrict the solver agent specifically associated with the role of "\textit{Programmer who can write/create computer software or applications with extensive computing and coding experience ...}" since this is the only appropriate role for writing solution code.

\begin{table}
\scalebox{0.85}{
    \begin{tabular}{@{}cc@{}}
    \toprule
    \textbf{Dimension Name} & \textbf{Dimension Definition} \\ \midrule
    Functionality & Is the code able to achieve all the goals specified in the task description? \\
    Robustness & Is the code snippet able to handle different unexpected inputs or other exceptions? \\
    Test Case Coverage & Does the solution contain test cases to cover all the software solution code? \\
    Documentation & Does the solution code contain enough comments or docstrings to explain itself? \\
    Code Style Violation & Does the solution code follow code style conventions defined in PEP 8? \\ \bottomrule
    \end{tabular}
}
\caption{concrete definitions of all evaluation dimensions we study in our work.}
\label{tab:dimension}
\end{table}

\section{Prompt for Model-Based Evaluation}
\label{sec:eval_prompt}
We show the evaluation prompts for generating scalar scores and feedback for functionality, robustness, and test case coverage dimensions in Figure \ref{fig:eval_prompt}, and for generating feedback for documentation dimension in Figure \ref{fig:eval_prompt_documentation}.
Critical components in the evaluation prompt include the evaluation dimension's name and definition, software task description, and software solution code. We list concrete definitions of all evaluation dimensions we study in Table \ref{tab:dimension}.

\section{Are More Optimization Steps Needed?}
\label{sec:more_steps}
We investigate whether more steps of optimization are beneficial compared with the current default steps of 5. We choose functionality and code style violation dimensions under the online group optimization with one-pass prompting setting and optimize for 10 steps. We plot the optimization curve on training and development set in Figure \ref{fig:more_steps}. We observe that for both dimensions, the training and development optimization curves tend towards either a stable or declining trend. This shows more steps of optimization are not necessary in our case study.

\begin{figure}[!h]
    \centering
    \begin{tikzpicture}[scale=0.70]

        \begin{axis}[
            name=train_plot,
            title={\normalsize \texttt{Functionality ($\uparrow$)}},
            ylabel={Evaluation Scores},
            xmin=-2, xmax=102,
            ymin=5.9, ymax=8.6,
            % xtick distance=1,
            xtick={0,10,20,30,40,50,60,70,80,90,100},
            xticklabels={0\%,10\%,20\%,30\%,40\%,50\%,60\%,70\%,80\%,90\%,100\%},
            ytick distance=0.5,
            xmajorgrids=true, ymajorgrids=true,            
            height=130pt,
            width=210pt,
            every axis/.append style={
                font=\scriptsize,
                % line width=1pt,
                % tick style={line width=3pt}
            },
            every axis plot/.append style={very thick},
            every axis legend/.append style={
                at={(1,1)},
                anchor=north east,
                % nodes={scale=0.54},
                legend cell align={left}
            },
        ]  
            \addplot[color=BrickRed,mark=triangle*,]  
            coordinates {(0,6.9)(10,8)(20,7.7)(30,7.6)(40,8.3)(50,7.6)(60,7.4)(70,6.9)(80,7.3)(90,7.6)(100,7.3)};
            \addplot[color=Violet,mark=*,]  
            coordinates {(00,6.48)(10,6.93)(20,7.13)(30,7.27)(40,6.8)(50,7.53)(60,7.07)(70,7.2)(80,7.07)(90,6.73)(100,6.87)};
            \legend{train, dev}
        \end{axis}
    
        \begin{axis}[
            name=dev_plot,
            at={($(train_plot.south west)+(0,-38pt)$)}, anchor=north west,
            title={\normalsize \texttt{Violation ($\downarrow$)}},
            ylabel={Evaluation Scores},
            xmin=-2, xmax=102,
            % xtick distance=1,
            xtick={0,10,20,30,40,50,60,70,80,90,100},
            xticklabels={0\%,10\%,20\%,30\%,40\%,50\%,60\%,70\%,80\%,90\%,100\%},
            ymin=-0.1, ymax=12.1,
            xtick distance=1,
            ytick distance=2,
            xmajorgrids=true, ymajorgrids=true,            
            height=130pt,
            width=210pt,
            every axis/.append style={
                font=\scriptsize,
                % line width=1pt,
                % tick style={line width=3pt}
            },
            every axis plot/.append style={very thick},
            every axis legend/.append style={
                at={(1,1)},
                anchor=north east,
                % nodes={scale=0.54},
                legend cell align={left}
            },
        ]
            \addplot[color=BrickRed,mark=triangle*,]  
            coordinates {(0,7.3)(10,5.3)(20,11.1)(30,4.6)(40,4.9)(50,3.8)(60,3.4)(70,3.5)(80,2.9)(90,3.9)(100,2.9)};
            \addplot[color=Violet,mark=*,]  
            coordinates {(0,7)(10,7.07)(20,4.13)(30,4.57)(40,3.43)(50,4.27)(60,3.33)(70,4.73)(80,4.17)(90,2.3)(100,3.47)};
            \legend{train, dev}
        \end{axis}

    \end{tikzpicture}
    \caption{Curve analysis for doubling the training steps for functionality and code style violation evaluation dimensions under online group with one-pass promoting optimization.}
    \label{fig:more_steps}
\end{figure}

\begin{figure}
    \centering
    \begin{tcolorbox}[fit basedim=10pt,left=5pt,right=5pt,colback=white,colframe=black,boxrule=1pt,fontupper=\ttfamily]
        {\tcbfontsize{1}
            \textbf{System Prompt}: \\
            
            \textbf{User Prompt}: \\
            You are faced with a software engineer task: \\
            \$\{task description\} \\
    
            You are also given a pool of experts: 
            
            [Experts pool] \\
            1. Chief executive officer whose main responsibilities include being an active decision-maker on users' demands and other key policy issues, leader, manager, and executor. \\
            2. Chief product officer who is responsible for all product-related matters, including product design, product strategy, product vision, product innovation, project management, and product marketing. \\
            3. Counselor whose main responsibilities include asking what users and customers think and providing valuable suggestions. \\
            4. Chief technology officer who is very familiar with information technology and will make high-level decisions for the overarching technology infrastructure that closely aligns with the organization's goals. \\
            5. Chief human resource officer who oversees all aspects of human resource management and industrial relations policies, practices and operations for an organization. \\
            6. Programmer who can write/create computer software or applications with extensive computing and coding experience in many varieties of programming languages and platforms, such as Python, Java, C, C++, HTML, CSS, JavaScript, XML, SQL, PHP, etc. \\
            7. Code reviewer who can help programmers assess source codes for software troubleshooting, fix bugs to increase code quality and robustness, and offer proposals to improve the source codes. \\
            8. Software test engineer who can use the software as intended to analyze its functional properties, design manual and automated test procedures to evaluate each software product, build and implement software evaluation test programs, and run test programs to ensure that testing protocols evaluate the software correctly. \\
            9. Chief creative officer who directs the company's creative software and develops the artistic design strategy that defines the company's brand.
            
            [/Experts pool] \\
    
            You need to recruit a total of \$\{number of agents\} experts from the above expert pool to collaboratively solve the given task. The first expert member you recruit must always be the programmer (index 6) since the programmer is responsible for developing the software code. The remaining expert members are responsible for providing feedback on the software code developed by the programmer. You can only select each expert once. \\
    
            Please use a comma to separate selected expert member indices without space. Always put the programmer at the beginning. Don't provide any reason for your selection. For example, if you recruit programmer, chief technology officer, and chief creative officer, you should output: \\
            6,4,9
        }
    \end{tcolorbox}
\caption{System and user prompt we use for the role selection agent for selecting concrete roles for solver and reviewer agents.}
\label{fig:recruiter_prompt}
\end{figure}

\begin{figure*}
    \centering
    \begin{tcolorbox}[left=5pt,right=5pt,colback=white,colframe=black,boxrule=1pt,fontupper=\ttfamily]
        \textbf{System Prompt}: \\
        You are a professional and strict code reviewer. \\ \\
        \textbf{User Prompt}: \\
        You will evaluate the solution code to a software engineer task from the following dimensions: \\
        
        \$\{dimension name\}: \$\{dimension definition\} \\ 
        
        The task description is: \\
        \$\{task description\} \\

        The solution code is:
        
        [Solution code] \\
        \$\{solution code\} 
        
        [/Solution code] \\

        Now, you need to give a rating from 0 to 10 with detailed reasons for your rating regarding the evaluation dimensions. \\

        You must only output one line which contains the score and detailed reasons why you gave this score. You must put detailed reasons inside <Reasons> tag. The exact output format you need to follow is shown below: \\
        1. \$\{dimension\}: a score (from 0 to 10) + <Reasons> detailed reasons </Reasons> \\

        Make sure your score and explanations of the score are only based on the evaluation dimension, not anything else. Also, make sure that you only give a high score when the solution code is really good at satisfying the definition of the evaluation dimension.
    \end{tcolorbox}
\caption{Evaluation system and user prompt for generating utility scores and textual feedback for functionality, robustness, and test case coverage dimensions.}
\label{fig:eval_prompt}
\end{figure*}

\begin{figure*}
    \centering
    \begin{tcolorbox}[left=5pt,right=5pt,colback=white,colframe=black,boxrule=1pt,fontupper=\ttfamily]
        \textbf{System Prompt}: \\
        You are a professional and strict code reviewer. \\ \\
        \textbf{User Prompt}: \\
        You will evaluate the solution code to a software engineer task from the following criterion: \\
        
        A good solution code always comes with abundant comments and docstrings to explain the purpose and functionality of each class, method, and function. \\ 
        
        The task description is: \\
        \$\{task description\} \\

        The solution code is:
        
        [Solution code] \\
        \$\{solution code\} 
        
        [/Solution code] \\

        You must use detailed natural language to describe how the solution code performs in terms of the above evaluation criterion. Make sure your evaluation is only related to the above evaluation criterion, nothing else.
    \end{tcolorbox}
\caption{Evaluation system and user prompt for generating textual feedback for documentation dimension.}
\label{fig:eval_prompt_documentation}
\end{figure*}

\begin{figure*}
    \centering
    \begin{tcolorbox}[fit basedim=8.5pt,left=5pt,right=5pt,colback=white,colframe=black,boxrule=1pt,fontupper=\ttfamily]
        {\tcbfontsize{1}
            \textbf{System Prompt}: \\
            You are a professional error agent locator who can accurately identify agents not functioning well in a multi-agent collaboration system. \\ \\
            \textbf{User Prompt}: \\
            You are given a software engineer task description, a group of agents that collaboratively solve this task with their communication trajectory, and accurate external feedback to the **final** solution code regarding some evaluation dimensions: \\ \\
    
            [Task description] \\
            \$\{task description\} 
            
            [/Task description] \\ \\
    
            [Agents with their role descriptions] \\
            Agent 1: \$\{agent 1 role description\} \\
            Agent 2: \$\{agent 2 role description\} \\
            Agent 3: \$\{agent 3 role description\} \\
            ... \\
            \$\{high-level responsibilities of each agents\} 
            
            [/Agents with their role descriptions] \\ \\
    
            [Agents communication trajectory] \\
            \$\{communication trajectory\}
            
            [/Agents communication trajectory] \\ \\
    
            [Feedback to final solution code] \\
            \$\{evaluation score with feedback\} 
            
            [/Feedback to final solution code] \\ \\ 
    
            You can assume all agents can possibly make mistakes when writing solution code or providing feedback to code, but the external feedback to the final solution is objective and robust.  \\ 
    
            If the above external feedback to the final solution code contains negative feedback, please identify all agent(s) causing the negative feedback and provide detailed explanations of why each identified agent leads to the negative feedback. Explanations must satisfy the following criteria: \\
            - If the identified agent improves itself based on the explanations, the negative feedback can be somehow mitigated, thus increasing the evaluation dimension score according to the definition of the evaluation dimension. \\
            - Make sure the explanations are specific and detailed instead of just general explanations. \\
            - Do not simply use 'agent 1' or 'agent 2' to describe agents in the explanations; instead, use their role descriptions, such as programmer and chief product officer, to describe. \\
            - Explanations should be based on what already happened in the trajectory instead of asking for more interactions between agents in the future. \\

            First, output the identified agent index, then provide detailed explanations. For example, if two agents, the programmer agent (Agent 1) and the code reviewer agent (Agent 3), are identified, the output must be two lines following the format below: \\
            1. Agent 1: detailed explanations of why programmer leads to negative feedback \\
            2. Agent 3: detailed explanations of why code reviewer leads to negative feedback \\

            If the feedback is completely positive, you can return a string of "None", meaning no agent is making mistakes.
        }
    \end{tcolorbox}
\caption{System and user prompt for the locator.}
\label{fig:locator_prompt}
\end{figure*}

\begin{figure*}
    \centering
    \begin{tcolorbox}[fit basedim=10pt,left=5pt,right=5pt,colback=white,colframe=black,boxrule=1pt,fontupper=\ttfamily]
        {\tcbfontsize{1}
            \textbf{System Prompt}: \\
            You are a professional system prompt optimizer for large language model agents. Don't be afraid to be creative. \\
            
            \textbf{User Prompt}: \\
            Below are the input system message, input user messages, and output of a large language model agent: \\ \\
            
            [Agent system message] \\
            \$\{agent system prompt\} 

            [/Agent system message] \\ \\

            [Agent user messages] \\
            \$\{agent user prompts\}
            
            [/Agent user messages] \\ \\ 

            [Agent output] \\
            \$\{agent output\}
            
            [/Agent output] \\ \\

            Below is the feedback for the above large language model agent activity trajectory: \\ \\ 

            [Fine-grained feedback] \\
            \$\{agent fine-grained feedback\} 
            
            [/Fine-grained feedback] \\ \\

            Based on the above feedback for this large language model agent, please ONLY update the system prompt wrapped between the <TO IMPROVE> tag in the agent system message so that the agent can improve based on the feedback.  \\

            You need to carefully think about what is useful in the feedback for you to optimize the agent system prompt and make sure that this optimization not only benefits the current software development task but can also generally benefit other software development tasks in the future. You must keep the new prompt clear, concise, informative, and descriptive. Also, make sure not to change the agent's original goal. \\ 

            Since your output will directly replace the text wrapped between the <TO IMPROVE> tag in the system prompt, make sure you ONLY output the improved prompt. DO NOT output anything else, such as the <TO IMPROVE> tag.
        }
    \end{tcolorbox}
\caption{System and user prompt for the optimizer.}
\label{fig:optimizer_prompt}
\end{figure*}

\begin{table*}[!t]
    \centering
    \scalebox{0.95}{
        \begin{tabular}{p{1\linewidth}}
             \toprule
             \textbf{Functionality} \\
             \midrule
             As a Software Test Engineer, your task is to provide detailed and specific feedback on the most recent solution code to the software task given. {\color{ForestGreen} \textbf{Your feedback should focus on the functionality, security, and versatility of the software.}} Test the software with different sports, player positions, and strategies, and with different user inputs. Also, test the user interface to ensure it is user-friendly and intuitive. {\color{ForestGreen} \textbf{Your feedback will guide the programmer in improving the solution and ensuring it meets the requirements of a wide range of users. Additionally, ensure your testing procedures cover all aspects of the task description and user experience, and provide more detailed feedback on any missing functionalities or areas for improvement}} After providing feedback, ensure that the programmer has understood and incorporated your suggestions into the final solution. \\
             \midrule
             \textbf{Robustness} \\
             \midrule
             As a Software Test Engineer, your task is to provide detailed and specific feedback on the most recent solution code to the software task given. {\color{ForestGreen} \textbf{Your feedback should focus on potential areas of robustness that may have been overlooked, such as error handling, input validation, and handling of unexpected exceptions.}} Also, consider the lifecycle of any temporary files created during the process and ensure they are properly managed to prevent unnecessary storage usage. {\color{ForestGreen} \textbf{Your feedback should help to ensure that the final solution is as robust and reliable as possible. You should have a deeper understanding of the code and the potential edge cases and error scenarios, and provide more detailed and specific feedback to the programmer agent.}} Furthermore, consider the integrity and validity of the data being processed, and suggest improvements to the data handling processes, including checks for invalid or duplicate inputs, and robust handling of data storage and retrieval. \\
             \midrule
             \textbf{Violation} \\
             \midrule
             As a Software Test Engineer, your role is to provide specific and detailed feedback on the most recent solution code to the software task given. While focusing on the functionality and robustness of the software, also consider the readability and maintainability of the code. {\color{ForestGreen} \textbf{Review the code for adherence to style guides like PEP8, including line length, and provide feedback on this aspect. If you notice any violations, point them out and suggest possible corrections.}} Also, consider the feedback from other teammates and reinforce any important points they might have raised. \\
             \bottomrule 
        \end{tabular}
    }
    \caption{We list optimized system prompts for the reviewer agent whose role is described as a software test engineer along functionality, robustness, and code violation evaluation dimensions. Green text indicates agent prompts are being optimized under the evaluation dimension.
}
    \label{tab:test_engineer_agent_case}
\end{table*}

\end{document}